\documentclass[review]{elsarticle}
\usepackage{amssymb}

\usepackage{hyperref}

\journal{}

\begin{document}

\begin{frontmatter}

\title{Adaptive Dependency Learning Graph Neural Networks}

\author[monash1]{Abishek~Sriramulu \corref{cor1}}
\author[unifrance]{Nicolas~Fourrier}
\author[monash1]{Christoph~Bergmeir}

\address{abishek.sriramulu@monash.edu, nfourrier@gmail.com, christoph.bergmeir@monash.edu}
\address[monash1]{Department of Data Science \& Artificial Intelligence, Monash University, Clayton VIC 3800, Australia.}
\address[unifrance]{Léonard de Vinci Pôle Universitaire, Research Center, 92 916 Paris La Défense, France.}

\cortext[cor1]{Postal Address: Faculty of Information Technology, P.O. Box 63 Monash University, Victoria 3800, Australia. E-mail address: abishek.sriramulu@monash.edu}

\begin{abstract}

Graph Neural Networks (GNN) have recently gained popularity in the forecasting domain due to their ability to model complex spatial and temporal patterns in tasks such as traffic forecasting and region-based demand forecasting. Most of these methods require a predefined graph as input,  whereas in real-life multivariate time series problems, a well-predefined dependency graph rarely exists. This requirement makes it harder for GNNs to be utilised widely for multivariate forecasting problems in other domains such as retail or energy. In this paper, we propose a hybrid approach combining neural networks and statistical structure learning models to self-learn the dependencies and construct a dynamically changing dependency graph from multivariate data aiming to enable the use of GNNs for multivariate forecasting even when a well-defined graph does not exist.
The statistical structure modeling in conjunction with neural networks provides a well-principled and efficient approach by bringing in causal semantics to determine dependencies among the series. Finally, we demonstrate significantly improved performance using our proposed approach on real-world benchmark datasets without a pre-defined dependency graph.
\end{abstract}


\end{frontmatter}

\section{Introduction}

In the present day, various communities in the world are positively impacted by understanding and modeling the series of data recorded over time such as the price of products in the retail industry, energy consumption, and movement of traffic. Frequently, multiple series of data are observed together at the same point in time and these series might exhibit relationships among themselves. Recent studies on Global Forecasting Models (GFM) have proven that many forecasting problems benefit hugely by generalising over multiple (related) time series, rather than modeling them isolated \cite{bandara2021improving}. Even though GFMs model global information across multiple time series, they assume that each series is recorded in isolation and that there exists no direct influence among them.

In situations with multivariate data, considering the fact that all the series have been recorded together at the same point in time can provide advantages while modeling the data. Modeling these inter-dependencies from multivariate data is significant but difficult as they are often complex and dynamic in nature. For example, an increase in the sales of one product may cause an increase or decrease in the sales of another product in a retail dataset due to several reasons, and this relationship may change over time.

Traditionally forecasts are made based on the historical values of the series. The Gaussian Process and different vector autoregression (VAR) models are a few of the most widely used forecasting methods for multivariate time series  \cite{Zivot2003VectorAM}. These models work under the assumption that there exists a linear relationship between every variable to every other variable in the system, which makes a direct interpretation of the estimated coefficients difficult. Furthermore, these models tend to over-fit while handling large numbers of variables \cite{wu2020connecting}.

The ability to capture complex relationships makes neural network models a successful alternative to solve multivariate forecasting problems \cite{lai_modeling_2018, guo_exploring_2019, dual_self_attn_2019}. Lai et al.~\cite{lai_modeling_2018} introduced a new variant of the recurrent neural network model called LSTNet which encodes the locality in series using 1D convolutional layers to a low dimensional representation. Then, the network decodes the output of the encoder using recurrent neural network blocks, however, the interdependencies between the series are not learned distinctly in LSTNet. 

Attention mechanisms have gained popularity in the machine learning space due to their ability to model dependencies without regard to their distance in the input or output sequences \cite{Bahdanau2015NeuralMT}. For time series forecasting, attention mechanisms have proven to improve modeling when dealing with longer sequences and also aid in improvising feature selection \cite{lim_temporal_2019}. The IMV-LSTM model proposed by Guo et al. \cite{guo_exploring_2019} has the ability to exploit the structure of Long Short-Term Memory (LSTM) to learn both the inter-series and intra-series dependencies simultaneously. The attention-based LSTM used in IMV-LSTM aids in learning the inner dynamics of the series. IMV-LSTM is observed to perform the best when the number of series is small but does not scale well when the series are large in number. Huang et al. \cite{dual_self_attn_2019} proposed DSANet which combines an autoregressive component with a dual self-
attention network where each of the series in a multivariate time
series is modeled independently in two convolutional components simultaneously to capture both the global and local temporal patterns in the data. These learned time series representations from the convolutional components are passed into individual attention modules to capture the inner dynamics of the series. 

\begin{figure*}[htb]
  \begin{center}
  \includegraphics[width=0.5\textwidth]{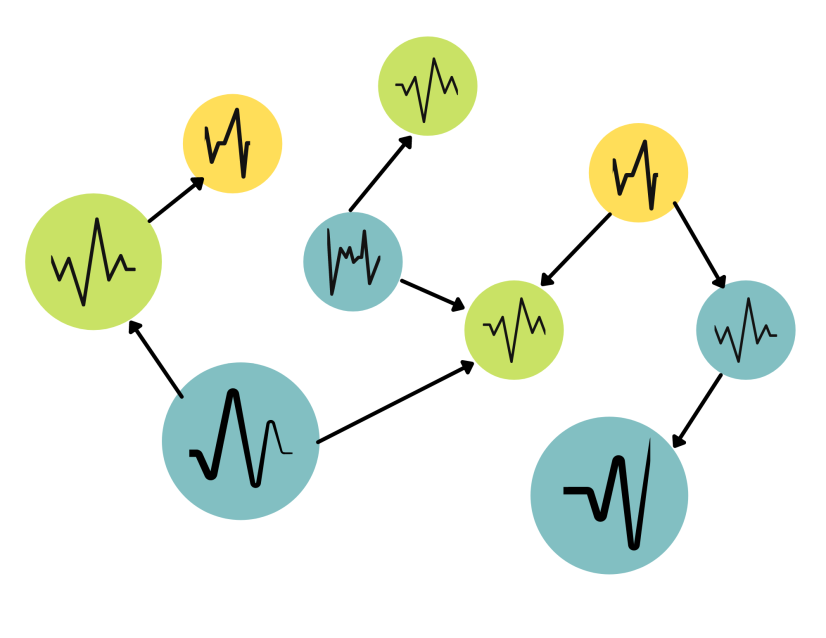}\\
  \caption{Multivariate data represented in the form of a graph.}\label{Graph_Example}
  \end{center}
\end{figure*}

Graph Neural Networks (GNN) have shown state-of-the-art performance in multivariate forecasting problems where a known graph structure is available such as traffic forecasting \cite{jiang2021graph}. GNNs are preferred in  settings such as traffic forecasting or financial time series where hierarchical or inter-series dependencies have significant influence \cite{chen2021novel}. In the context of traffic forecasting, spatial dependency can naturally be represented through graphs making GNN an excellent candidate for modeling, where road intersections are the nodes and road connections the edges \cite{jiang2021graph}. However, many multivariate forecasting problems in the real world do not have a prior well-defined graph, which hinders the use of GNNs in these cases. To overcome this, Wu et al. \cite{wu2020connecting} propose constructing a graph structure to enable the use of GNNs in problems where a well-defined graph does not exist. Consider the graph shown in Fig.~\ref{Graph_Example}. Here, the nodes represent series in a multivariate time series dataset where the edges connecting them represent the relationships among the series. In this way, any multivariate time series data can be represented as graphs by estimating the relationship among the series. The limitation of the graph construction method presented in \cite{wu2020connecting} is that the complexity is high as the method involves training the model to learn the weights of two $N \times N$ matrices where $N$ is the number of series. To the best of our knowledge, the work of Wu et al. \cite{wu2020connecting} is the only neural network approach to constructing graph structures from multivariate time series data for time series forecasting application, which is the base of our research work.

In multivariate forecasting problems, it is important to model the intra-series temporal dependencies along with the inter-series spatial dependencies that exist within the series history. 
The following are the contributions of our work:
\begin{itemize}
\item We introduce the Adaptive Dependency Learning Neural Network (ADLNN) that has the ability to learn the dynamically changing dependencies among the series by bringing in inter-series causal semantics along with modeling the intra-series dependencies to improve the forecasting accuracy.
\item We introduce a well-principled approach for initialising the adjacency matrix when a pre-defined graph structure is not provided.
\item We introduce the Dynamic Graph Construction Block that converts the static adjacency matrix into a dynamic matrix that helps model the changes in the relationship between time series over time.
\item We study and compare the performance in terms of accuracy and time complexity of our methods against the existing state-of-art models on real-world datasets.
\end{itemize}

Some approaches in other areas that combine GNNs and statistical modeling have been proposed in the literature \cite{zhang2020efficient, qu2019probabilistic, jin2019graph, qu2019probabilistic, qu2019gmnn}. Our approach is to the best of our knowledge the first that combines Statistical modeling and neural network for graph construction in a multivariate time series setting.

In our proposed model, we initialise the adjacency matrix using various computationally efficient statistical methods which are then fed into a convolutional attention mechanism to dynamically weight the matrix in order to model the dynamic changes in dependency. We include alternating blocks of graph convolution and temporal convolution to model both the spatial and temporal patterns simultaneously. Furthermore, we combine a spatio-temporal convolutional attention forecasting model along with the GNN to improve the temporal modeling ability. Our code and models are made publicly available at https://github.com/AbishekSriramulu/ADLGNN.git.

\section{Related Works}
\subsection{Multivariate time series forecasting}
Some of the major challenges of multivariate time series forecasting are nonlinearity and complex dynamic inter-series and intra-series dependencies. The prominently used traditional methods for multivariate forecasting are vector autoregression (VAR) models and  Gaussian Processes (GP). The VAR model is a statistical model that captures linear dependencies among multiple variables that change over time \cite{Zivot2003VectorAM}. VAR models are linear autoregressive models, which are a general class of stochastic models used for forecasting time series which can be transformed into stationary series \cite{boxjen76}. In VAR, each series has an equation to model its evolution over time and this equation includes several lags of the series along with the lags of other series in the multivariate dataset. GPs are a type of Bayesian model used to model the multivariate distribution of inputs, so they can be easily used on multivariate data. However, these models tend to overfit when handling datasets with many variables. Moreover, they also involve high computational costs so they do not scale well. Many deep learning models have been proposed for the last few years to overcome these issues. Convolutional Neural Networks (CNN) and Recurrent Neural Networks (RNN) have been widely applied in time series forecasting to capture complex nonlinear dependencies. Some noteworthy models are long short-term memory (LSTM) \cite{hochreiter1997long}, LSTNet \cite{lai_modeling_2018} and Temporal Convolution Network (TCN) \cite{bai2018empirical}. However, these models fail to capture the inter-series relationships.

\begin{table}[!htb]
\caption{Notations used and their description}
\begin{tabular}{@{}ll@{}}
  Notations & Description\\
\hline

 $G$ & Multivariate graph \\
  $V$ & Set of nodes in a graph \\
  $E$ & Set of edges in a graph \\
  $N$ & Number of nodes in a graph\\
   $A$ & Adjacency Matrix \\
   $A_{CM}$ & Correlation Matrix \\
   $A_{GC}$ & Granger Causality Matrix \\
   $A_{CST}$ & Correlation Spanning Tree Matrix \\
   $A_{GL}$ & Graphical Lasso Matrix \\
   $A_{MLE}$ & Maximum Likelihood Estimation Matrix \\
   $A_{MI}$ & Mutual Information Matrix \\
   $A_{TE}$ & Transfer Entropy Matrix \\
   $Y$ & Historical Multivariate Time Series variable \\
   $\hat Y$ & Future sequence of Y \\
   $argtopS$ & Function that returns the top S values \\
\hline
\end{tabular}
\end{table}

\subsection{Graph Neural Networks}
\label{adh_mat_expl}
A graph is a data structure that defines dependencies between objects or entities. A graph is mathematically represented as $ G = (V,E)$ where $V$ represents all nodes, and all edges are in the set $E$. 
The total number of nodes in the graph is denoted by $n$. 
Node $v\in V$ is said to be in the neighborhood of node $u\in V$ in a graph $G$ if there exists an edge from node $v$ to $u$ or vice versa. 
A graph structure can be defined as an adjacency matrix. Adjacency matrices are mathematically denoted as \( A \in \mathbb{R}^{n \times n} \) where \( A[i,j] > 0 \) if \( (v_{i}, v_{j}) \in E \) and \( A[i,j] = 0 \) if \( (v_{i}, v_{j}) \notin E \) \cite{wu2020connecting}.

Graph neural networks have been successfully used in many machine learning applications involving graphs due to their ability to directly handle graph data.
GNNs are based on a concept of information propagation, nodes in a graph exchange information among their neighbors \cite{klicpera2018combining}. The state of each node in a graph depends on the state of its neighbors. This ability of GNNs allows the model to capture spatial dependencies in a network \cite{wu2020connecting}. Convolutional GNNs, graph autoencoders, recurrent GNNs, and spatio-temporal GNNs are the most widely used variants of GNN \cite{gnn_review_2019}. %
Spatio-temporal GNNs are more suitable for multivariate time series forecasting due to their ability to capture the temporal patterns in the spatial domain. In spatio-temporal GNNs, the graph convolution module models the spatial dependencies while 1D convolution layers model the temporal patterns in the data \cite{yan2018spatial}. The graph structure that exists in the data needs to be explicitly represented in an adjacency matrix which can be fed into the graph convolutions as input. 
Dynamic graphs were used in \cite{peng2020spatial}, with external time-based features to change the traffic flow graph in a pre-specified way. To introduce this dynamicity in the graph Huang et al.~\cite{huang2022gan} introduced a weighted matrix based on irregularities in the relationships between regions. Even though these methods can be effective in the specific applications they are developed for, they cannot easily be transferred to other applications. The use of an attention mechanism to introduce dynamicity allows the dynamic graph approach to be scaled to any application but at $O(N^2)$ complexity, where $N$ is the number of nodes \cite{ali2021exploiting, sun2022sequential, li2020dynamic}. 

\subsection{Graph Structure Construction}
\label{static_graph_expl}
Multivariate time series data are arguably best represented as graphs. An increased understanding of the structural characteristics of the data enables better analysis and prediction of such systems. However, most of the multivariate data observed in nature do not exist with an explicit graph structure. In such cases, network science concepts can be used to infer the underlying graph structure from the data.
The most widely used graph structure construction methods are the Bayesian network structure learning methods such as Grow-Shrink (GS), Incremental, Association Markov Blanket (IAMB), Fast Incremental Association (Fast-IAMB), Interleaved Incremental Association (Inter-IAMB), Incremental Association with FDR Correction (IAMB-FDR), Max-Min Parents and Children (MMPC), Semi-Interleaved Hiton-PC (SI-HITON-PC) and Hybrid Parents and Children (HPC) \cite{scutari2009learning}. However, these methods are not suitable for high dimensional data due to their high computation costs \cite{hartle2020network}. In \cite{li2018dynamic}, an approach involving the learning of affinity graph and feature fusion, for resulting in better clustering results was proposed. This method also assigns the affinity weights for data points on a per-data-pair basis to avoid the explicit specification of the size of the neighborhood.  In \cite{luo2017adaptive}, the similarity matrix is initialised according to the gaussian function, this matrix is updated with a network that performs semi-supervised joint feature selection. In \cite{zhou2019person}, The initial adjacency matrix is either a similarity matrix or a pre-defined matrix which is then added to a sparse constraint in order to model the evolving node-to-node associations in a graph. Moreover, the complexity of these methods is observed to be at $O(N^2)$ where $N$ is the number of nodes. In \cite{peng2020spatial, peng2021dynamic}, an incidence dynamic graph structure called a traffic flow probability graph is obtained from historical data by calculating the normalised probability score of travel from one station to another while modeling for traffic flow forecasting. In \cite{guan2022dynagraph}, an efficient dynamic GNN training framework that reduces the overhead by cross-layer optimizations across GNN and RNN operations is introduced. In \cite{you2022roland}, A multi-hop mechanism is used to capture node information from the neighborhood to update the embedding
states over time. This approach can easily enable any static GNNs to handle dynamic graphs. However, this also increases the number of parameters of the model.
Since in most real-life cases, we deal with large numbers of series and it is more practical to use cost-effective methods. Hence, in this paper we use the following simple concepts from information theory to infer dependencies among the series:
\begin{itemize}
\item \textbf{Correlation Matrix (CM): } The adjacency matrix is constructed by using the pairwise Pearson's correlation coefficient across the series.
\item \textbf{Granger Causality (GC): } The graph structure is inferred by the effect of one series over another. The effect of series $Y_{2}$ on series $Y_{1}$ can be mathematically formulated as $log(e_1/e_{12})$ where $e_1$ is the error from an autoregressive model trained on data of $Y_{1}$ and $e_{12}$ is the error from an autoregressive model trained on data from $Y_{1}$ and $Y_{2}$.
\item \textbf{Correlation Spanning Tree (CST): }The graph is constructed by building a minimum spanning tree connecting all the series in the data using the distance matrix computed from Pearson's correlation matrix \cite{mantegna1999hierarchical}.
\item \textbf{Graphical Lasso (GL): }The graph structure is constructed using the inverse covariance matrix of the multivariate data estimated using the graphical lasso method proposed by Friedman et al.~\cite{friedman2008sparse}
\item \textbf{Maximum Likelihood Reconstruction (MLE): }The graph structure is inferred using the coupling matrix obtained from the MLR method proposed by Zeng et al.~\cite{zeng2013maximum}.
\item \textbf{Mutual Information (MI): }The adjacency matrix is obtained by estimating the mutual information between the probability distributions of each pair of series in the data. Mutual information measures by how much knowledge of one variable reduces uncertainty about the other.
\item \textbf{Transfer Entropy (TE): } 
The adjacency matrix is obtained from the pairwise transfer entropy estimation of each series with the others in the dataset. Transfer entropy is a measure of information contained about the future states of one series by knowing the past states of that series along with the other series \cite{lizier2014jidt}.
\end{itemize}

\subsection{Attention Mechanism}
Transformer models have become quite popular in recent years to solve many problems in computer vision (CV), natural language processing (NLP), and in time series forecasting. One of the most fundamental blocks of these kinds of models is the attention mechanism. Originally, it was developed to solve problems in Seq2seq model architectures used for neural machine translation, which encode input to a representation form, and pass it to a decoder to get predictions \cite{sutskever2014sequence}. For learning larger sequences, one of the biggest problems with this kind of architecture is the first state of the decoder, which only takes the output of the last state of the encoder. This has a bottleneck problem as none of the previous encoder states are taken into consideration while decoder weights are learned. An attention mechanism can solve this problem by creating the context vector from all states of the encoder to pass it to the decoder. During context vector learning, the attention weights are learned and are multiplied with the output of each state of the encoder. These weights represent the contribution made by each state in generating a final representation to pass to the output.

Bahdanau et al.~\cite{Bahdanau2015NeuralMT} first introduced the additive attention mechanism for neural machine translation which allowed the model to align towards segments of a sentence which are useful to detect the target word. Luong et al.~\cite{luong-etal-2015-effective} proposed a global multiplicative attention concept which takes all hidden states of the encoder into consideration to compute the representation of the context vector. Xu et al.~\cite{pmlr-v37-xuc15} also proposed soft and hard attention mechanisms to focus on certain parts of the image for image caption generation. Soft attention calculates the weighted sum of each state of the encoder, while hard attention picks the encoder state having the highest weight to build the context vector. Vaswani et al.~\cite{vaswani_attention_17} revolutionized the NLP space by proposing scaled dot product attention and multi-head attention. They proposed an architecture that completely replaces recurrence and convolution blocks. This network is highly parallelizable, so it can be trained faster \cite{vaswani_attention_17}. Scaled dot product attention is comprised of the dot product of inputs, which are queries and keys. This dot product is multiplied with a values vector and the final output is passed through a softmax function. Parallel calculations of such attention functions form multi-head attention blocks where the output of each dot product attention is concatenated and passed through a linear transformation function to calculate the output.

\begin{figure*}[htb]
	\centering
	\includegraphics[width=0.8\textwidth]{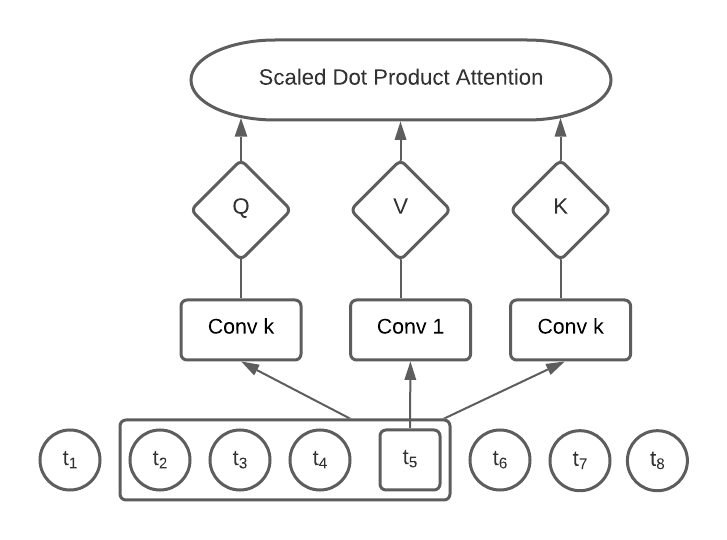}
	\caption{Architecture of convolutional self-attention}
	\label{fig:conv_attn_module}
\end{figure*}

Fan et al.~\cite{fan_multi_horizon_2019} introduce a neural network model that feeds univariate time series into an attention mechanism to capture temporal patterns. This model generates predictions for multiple horizons at once based on the latent variables. Lim et al.~\cite{lim_temporal_2019} introduced a new variant of the transformer model called temporal fusion transformer which uses self-attention layers to capture long-term dependencies. These models build attention on hidden states from the inputs, so the attention weights' direction of correlation with the target is not taken into consideration. All these variants of attention are insensitive to the local context as they use a point-wise dot product self-attention. To overcome this, Li et al. \cite{li2019enhancing} present convolutional self-attention where the queries and keys are passed through a causal convolution layer to incorporate the local context into the attention. Figure \ref{fig:conv_attn_module} shows the architecture of causal convolution, properly padded convolution of kernel size $k$ with stride 1 is used to transform the inputs into queries and keys. In causal convolutions, the similarities are computed by their local shapes, instead of point-wise values to provide the local context.

\section{Framework}

\subsection{Problem Formulation}
\label{headings}
Let $\mathrm{Y} \in \mathbb{R}^{T\times N}$ be a multivariate variable of dimension $N$ for times $1\ldots T$. Let $Y_{t} \in \mathbb{R}^{N}$ be the values of $\mathrm{Y}$ at time $t$, and $Y_{(i,t)} \in \mathbb{R}$ be the value of variable $i$ at time $t$. Thus, $\mathrm{Y} = \{Y_{1}, Y_{2}, ... ,Y_{T} \}$. Our goal is to predict the next values for the horizon of length $H$ for the $N$ variables. Hence, we predict the future values of sequence $\mathrm{Y}$ as $\mathrm{\hat{Y}} = \{\hat{Y}_{T+1}, \hat{Y}_{T+2}, \ldots ,\hat{Y}_{T+H} \}$. Our goal is to create a function that maps from $\mathrm{Y}$ to $\mathrm{\hat{Y}}$ such that loss is minimized with L2 regularization. 

\begin{figure}[!htb]
	\centering
	\includegraphics[width=0.9\textwidth]{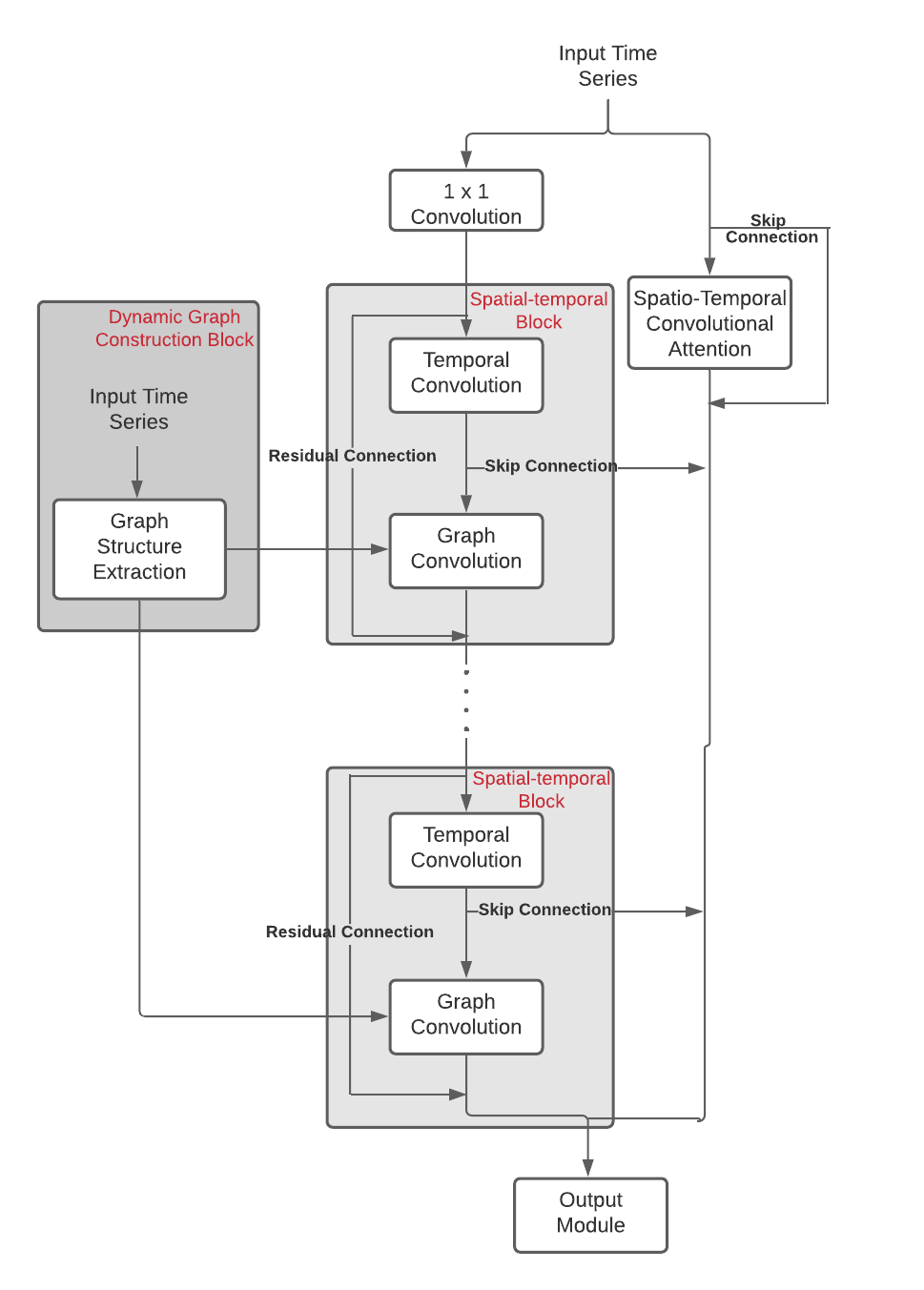}
	\caption{Model Architecture }
	\label{fig:model_arch}
\end{figure}

\begin{figure*}[htb]
	\centering
	\includegraphics[width=\textwidth]{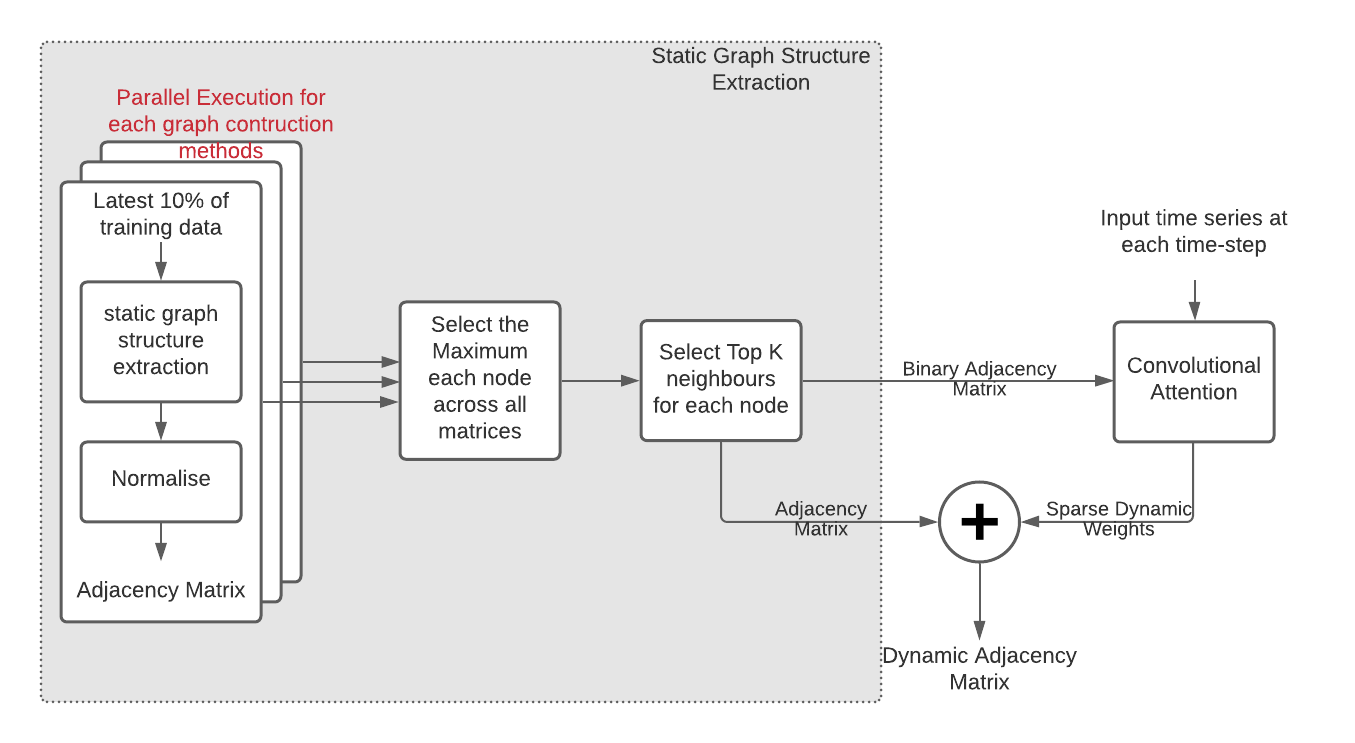}
	\caption{Dynamic Graph Construction Block}
	\label{fig:AdjMat}
\end{figure*}

\subsection{Model Architecture}

As shown in Figure \ref{fig:model_arch}, the model architecture consists of a dynamic graph construction block, several spatio-temporal blocks which contain a temporal convolution module and graph convolution module, a spatio-temporal convolutional attention and output module. The residual connections are included to avoid vanishing gradient problems and skip connections are used to avoid performance degradation due to the depth of the network.

\subsection{Dynamic graph construction block}

A multivariate time series is a complex system with inter-series dependencies that mostly change over time. To achieve a good performance, it is important to model these inter-series dynamics. A dynamic graph construction block takes in the multivariate time series as input and extracts the dependency graph that represents the inter-series dependencies at each time step. The output of this block is an adjacency matrix for each time step in the data, where the strength of relationships between series is specified with a value between 0 and 1 as mentioned in Section~\ref{adh_mat_expl}. The process of graph extraction is challenging for large and long datasets due to its complexity, so this block is carefully designed with well-principled techniques that require less computation. As shown in Figure \ref{fig:AdjMat}, firstly we extract adjacency matrices from the training data using the following methods:
Correlation Matrix, Granger Causality, Correlation Spanning Tree, Graphical Lasso
, Maximum Likelihood Estimation, Mutual Information, and Transfer Entropy (refer to Section~\ref{static_graph_expl} for details). Each of these methods outputs an adjacency matrix $A_{x}$ and these matrices are normalised to have values in the range 0 to 1. The resultant matrices are then reduced to a single matrix by selecting the maximum value of each cell across all the matrices as shown in Equation~\ref{equ:mat} where $S$ is a hyper-parameter that limits the neighborhood to the most influential neighbors. To relax the computation strain we only consider the latest 10\% of the training data to compute this static graph structure. Moreover, this initialised structure will also be updated as we train the network over the entire training set.

\begin{equation}
\label{equ:mat}
$$

$A(i,j) = \textit{max}_{x}(A_{x}(i,j)), \ \forall \ i,j \in \{1,2,...,N \}$

\qquad \qquad \qquad \qquad where $A_{x} \in \{ A_{CM}, A_{GC}, A_{CST}, A_{GL}, A_{MLE}, A_{MI}, A_{TE}  \} $

${A}[i,{k}] = 0,  \forall k \notin \textit{argtopS}(A[i,:]), \ \forall i \in \{1, 2, ...,N\}$

$$
\end{equation}

As mentioned in Figure \ref{fig:AdjMat}, a single adjacency matrix is obtained by selecting the maximum value of each cell across the matrices and then the top $K$ neighbors are selected for each node to create a sparse matrix in order to reduce the complexity of the model. From this reduced sparse adjacency matrix we create a binary adjacency matrix where $A[i,j] > 0$ is set to $1$.m   nhkppjkpjkk,kj.''
In Vaswani et al.~\cite{vaswani_attention_17} and Lim et al.~\cite{lim_temporal_2019}, the attention mechanism has been used to estimate the importance of a variable at a given time to predict the future state of a target variable. We use the same idea here to estimate the influence of one variable over the other at each time step. 

In order to make this static graph into a dynamic one, we use sparse convolutional attention which takes in the binary adjacency matrix $A_{Bi}$ along with the input time series at each time step as input to learn an $N \times K$ sparse dynamic weight matrix. These weights determine the strength of dependency of a series over the other. This weight is then summed with the static adjacency matrix obtained previously to construct a dynamic adjacency matrix. The static adjacency matrix provides a well-principled approach to reducing the complexity of the model from $O(N^2)$ to $O(N*K)$. Moreover, self-attention is used instead of multi-head attention to reduce the complexity of the model. The operation of convolutional attention in this block can be defined as in Equation~\ref{attn1}.

\begin{equation}
\label{attn1}
\textit{Attention}(q, k, v) = \textit{softmax}\left(\frac{qk^T}{\sqrt{d_{k}}} \cdot {A_{Bi}} \right)v
\end{equation}

Here, $d_{k}$ is the input time series that is transformed into query vector $q$, a key vector $k$, and a value vector $v$ using causal convolutions, and $A_{Bi}$ is the binary adjacency matrix that is used to mask the attention mechanism to reduce complexity.

\subsection{Spatio-temporal convolutional attention}

\begin{figure}[htb]
	\centering
	\includegraphics[width=0.8\textwidth]{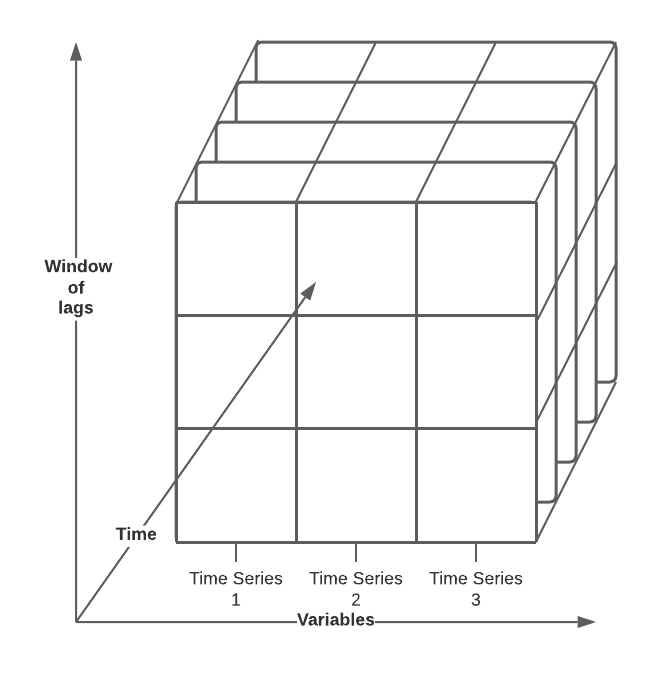}
	\caption{Input data shape}
	\label{fig:3dinput}
\end{figure}

Figure~\ref{fig:3dinput} shows the shape of the input data that is fed into the model. In a spatio-temporal convolutional attention, convolutional attention transforms the inputs at each time step into queries and keys. We use causal convolutions to ensure that at any point in time the future states are not accessible. The input shape of the data ensures that the model attends to both the time and spatial domains simultaneously. Moreover, using a large kernel size would increase the local context awareness of the model and reduce the impact of anomalies or change points. As shown in Figure~\ref{fig:conv_attn_module}, the input $d_{k}$ is transformed into query vector $q$, a key vector $k$, and a value vector $v$ of the same size. A score is estimated for each value in the input against each other. This learned attention score is then normalised using a softmax function, as in Equation~\ref{attn2}.

\begin{equation}
\label{attn2}
\textit{Attention}(q, k, v) = \textit{softmax}\left(\frac{qk^T}{\sqrt{d_{k}}} \right)v
\end{equation}

Equations~\ref{attn1} and \ref{attn2} are the same except for the masking. The spatio-temporal convolutional attention does not have any masks. The attention score is a measure of the influence each value in the input has on the other. This score is then fed into skip connections which are a standard convolution layer of shape $1 \times L_{W}$ where $L_{W}$ is the length of input windows. The skip connection standardises the length of intermediate layer outputs that are fed into the output module. 

\subsection{Graph Convolution Module}
\begin{figure}[htb]
	\centering
	\includegraphics[width= \linewidth]{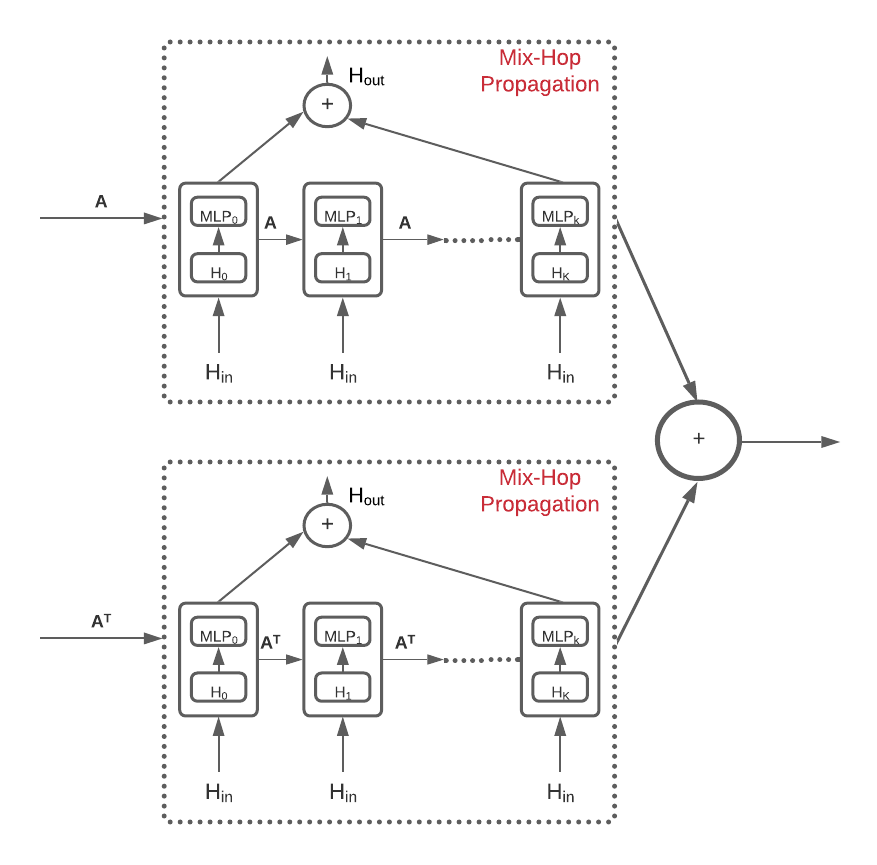}
	\caption{Graph Convolution Module}
	\label{fig:gcn}
\end{figure}
The graph convolution module that we utilise in this model is based on the work of Wu et al. \cite{wu2020connecting}. This module handles two main operations - information propagation and information selection. The adjacency matrix is fed into this module as input and the graph convolutions use the spatial information obtained from the adjacency matrix to propagate the information over the graph structure and select the information. This information flow is handled by two mix-hop propagation modules as shown in Figure~\ref{fig:gcn}. When using many graph convolution layers, the propagation of information happens recursively and this causes the node's hidden states to converge to a single point. Studies have recorded that when using many layers, the hidden states of nodes reach a random walk distribution \cite{klicpera2018predict}. To overcome this, we retain a fraction of the node's initial states. The information propagation operation in the mix-hop propagation module is defined as shown in Equations~\ref{info_prop} and \ref{info_selection}.
\begin{equation}
\label{info_prop}
H_{k} = \beta H_{in} + (1 - \beta) \hat{A} H_{k-1}                     
\end{equation}
\begin{equation}
\label{info_selection}
H_{out} = \beta H_{k} W_{k}                    
\end{equation}
Here, $\beta$ is a hyper-parameter that controls how much of the initial state to retain, $k$ is the depth of propagation, $H_{in}$ represents the initial hidden states, and $H_{out}$ represents the output hidden states. Equation~\ref{info_prop} represents how the information propagates horizontally across nodes and Equation~\ref{info_selection} shows how the MLP attaches a learnable parameter matrix to the current hidden state that selects the important information and disregards noise. 

\subsection{Temporal convolution module}

In order to capture the intra-series temporal patterns, we employ a temporal convolution module. we choose temporal convolutions over LSTM or RNN architectures since they do not backpropagate over the temporal dimension of the sequence. This avoids exploding or vanishing gradient problems. Moreover, temporal convolutions have proven to perform well on longer sequences \cite{dual_self_attn_2019}. Since temporal convolutions use causal convolution, information leakage is prevented as a state can only access the previous states and not the future. In our architecture, the temporal convolution module consists of multiple causal convolution filters of different sizes with dilation. This allows the module to extract patterns at various ranges and improves the ability to model long sequences. This removes the need to choose the right kernel size for convolutional networks. We use filter sizes of $1 \times 2$, $1 \times 3$, $1 \times 6$, and $1 \times 7$, to capture both short-term and long-term patterns. The outputs of these four convolution filters are first truncated to the length of the largest filter and concatenated over the channel dimension. For $z\in \mathbb{R}^{T}$ as a 1D input sequence, and 4 causal convolutional filters of sizes $\mathrm{f_{1\times2}}$, $\mathrm{f_{1\times3}}$, $\mathrm{f_{1\times6}}$, and $\mathrm{f_{1\times7}}$, the output of the temporal convolution module is given as shown in Equation~\ref{equ:convout}, where $z\star f_{1\times k}$ denotes the dilated convolution, defined in Equation~\ref{equ:dilconv}, and $d$ denotes the factor for dilation.

\begin{equation}
\label{equ:convout}
\mathrm{z} = \textit{concat}\mathrm{(z\star f_{1\times2}, z\star f_{1\times3}, z\star f_{1\times6}, z\star f_{1\times7})}
\end{equation}

\begin{equation}
\label{equ:dilconv}
\mathrm{z\star f_{1\times k}}(t) = \sum_{s=0}^{k-1}\mathrm{f_{1\times k}}(s)\mathrm{z}(t-d\times s)
\end{equation}

\subsection{Output Module}

The output module is used to transform the channel dimension of the current state to output dimensions. To achieve this, we use two convolution layers of size $1 \times 1$.

\section{Experiments and results}

\subsection{Datasets and performance metrics}

We use three benchmark datasets for all the experiments to evaluate the performance of the model against the state-of-the-art and baseline models. These benchmark datasets were used by Lai et al.~\cite{lai_modeling_2018} and Wu et al.~\cite{wu2020connecting} which allows us to make straightforward comparisons with those works. Information about dataset statistics is provided in Table~\ref{tab:table1}.

\begin{table}[!htb]
	\begin{center}
		\caption{Multivariate time series dataset Statistics}
		\label{tab:table1}
		\begin{tabular}{lccr}

			\hline
			\textbf{Datasets} & \textbf{Series lengths} & \textbf{Total series/nodes} & \textbf{Sample rate}\\
			\hline

			Solar energy\footnote{http://www.nrel.gov/grid/solar-power-data.html} & 52,560 & 137 & 10 minutes\\
			Electricity\footnote{https://archive.ics.uci.edu/ml/datasets/ElectricityLoadDiagrams20112014} & 26,304 & 321 & 1 hour\\
			Traffic\footnote{http://pems.dot.ca.gov} & 17,544 & 862 & 1 hour\\
			\hline

		\end{tabular}
	\end{center}
\end{table}

\paragraph{Solar energy dataset} This dataset comprises of solar energy production data from a solar power plant collected in 2006. These data were logged every 10 minutes from different power plants situated in the state of Alabama, USA \cite{lai_modeling_2018}. 

\paragraph{Electricity dataset} This dataset consists of electricity consumption records measured in kWh from 2012 to 2014 for 321 clients. We use the hourly data converted from the original data which was logged every 15 minutes \cite{lai_modeling_2018}.     

\paragraph{Traffic dataset} This dataset is a collection of 48 months of data from the Department of Transportation of California collected in the period from 2015 to 2016 on an hourly basis. The dataset contains the log of the road occupancy rates, which is estimated by multiple sensors on freeways located in the San Francisco Bay area.

For performance evaluation, following Wu et al.~\cite{wu2020connecting}, we use the following evaluation metrics. 

\textbf{Root Relative Squared Error (RSE):} This metric is a modified scaled variant of the popular Root Mean Squared Error (RMSE), which is developed to ensure a meaningful evaluation, irrespective of the size of the data. As per Lai et al.~\cite{lai_modeling_2018}, it is defined as in Equation~\ref{equ:RSE}.
	\begin{equation}
	\label{equ:RSE}
	\mathrm{RSE} = \frac{\sqrt{\sum_{(i, t)\in\Omega_{Test}}{{(Y_{(i,t)}-\hat{Y}_{(i,t)})}^2}}}{\sqrt{\sum_{(i, t)\in\Omega_{Test}}{{(Y_{(i,t)}-mean(Y))}^2}}}
	\end{equation}
	
\textbf{Correlation Coefficient (CORR):} It is an estimate of the intensity of the relation between the relative variations between two variables. As per Lai et al.~\cite{lai_modeling_2018}, the formula of CORR is given as in Equation~\ref{equ:CORR}.
	\begin{equation}
	\label{equ:CORR}
	\mathrm{CORR} = \frac{1}{n}\sum_{i=1}^n\frac{\sum_{t}{(Y_{(i,t)}-mean(Y_{(i,)}))(\hat{Y}_{(i,t)}-mean(\hat{Y}_{(i,)}))}}{\sqrt{\sum_{t}{{(Y_{(i,t)}-mean(Y_{(i,)}))}^2{(\hat{Y}_{(i,t)}-mean(\hat{Y}_{(i,)}))}^2}}}
	\end{equation}

Here, $Y, \hat{Y} \in {\mathbb{R}^{n\times{T}}}$, where $Y$ denotes ground truth values and $\hat{Y}$ denotes the values predicted by the model. For performance, lower values of RSE are better, while for CORR, higher values are better. While these measures may not be generally applicable in a forecasting context due to non-stationarities (e.g., if there is a strong trend in the series, calculating a mean is essentially meaningless), there doesn't exist a general consensus about how to evaluate forecasts, and these measures are applicable to our datasets and make our results comparable to results published in the literature \cite{kolassa2016evaluating}.

\begin{table}[!htb]
	\caption{Comparison of state-of-the-art methods for multivariate single-step time series forecasting }
	\label{results-table}
	\centering
	\resizebox{\textwidth}{!}{ 
		\begin{tabular}{@{}ll|cccc|cccc|cccc@{}}
			\hline
			\multicolumn{2}{c|}{Dataset} & \multicolumn{4}{c|}{Solar energy} & \multicolumn{4}{c|}{Electricity} & \multicolumn{4}{c}{Traffic} \\ \hline			\multicolumn{2}{l|}{} & \multicolumn{4}{c|}{Horizon} & \multicolumn{4}{c|}{Horizon} & \multicolumn{4}{c}{Horizon} \\ \hline
			Methods & Metric & 3 & 6 & 12 & 24 & 3 & 6 & 12 & 24 & 3 & 6 & 12 & 24 \\ 
			\hline
			AR & RSE & 0.2435 & 0.3790 & 0.5911 & 0.8699 & 0.0995 & 0.1035 & 0.1050 & 0.1054 & 0.5911 & 0.6218 & 0.6252 & 0.6300 \\
			& CORR & 0.9710 & 0.9263 & 0.8107 & 0.5314 & 0.8845 & 0.8632 & 0.8591 & 0.8595 & 0.7752 & 0.7568 & 0.7544 & 0.7519 \\ 
			\hline
			VAR-MLP & RSE & 0.1922 & 0.2679 & 0.4244 & 0.6841 & 0.1393 & 0.1620 & 0.1557 & 0.1274 & 0.5582 & 0.6579 & 0.6023 & 0.6146 \\
			& CORR & 0.9829 & 0.9655 & 0.9058 & 0.7149 & 0.8708 & 0.8389 & 0.8192 & 0.8679 & 0.8245 & 0.7695 & 0.7929 & 0.7891 \\ 
			\hline
			GP & RSE & 0.2259 & 0.3286 & 0.5200 & 0.7973 & 0.1500 & 0.1907 & 0.1621 & 0.1273 & 0.6082 & 0.6772 & 0.6406 & 0.5995 \\
			& CORR & 0.9751 & 0.9448 & 0.8518 & 0.5971 & 0.8670 & 0.8334 & 0.8394 & 0.8818 & 0.7831 & 0.7406 & 0.7671 & 0.7909 \\ 
			\hline
			RNN-GRU & RSE & 0.1932 & 0.2628 & 0.4163 & 0.4852 & 0.1102 & 0.1144 & 0.1183 & 0.1295 & 0.5358 & 0.5522 & 0.5562 & 0.5633 \\
			& CORR & 0.9823 & 0.9675 & 0.9150 & 0.8823 & 0.8597 & 0.8623 & 0.8472 & 0.8651 & 0.8511 & 0.8405 & 0.8345 & 0.8300 \\ 
			\hline
			LSTNet & RSE & 0.1843 & 0.2559 & 0.3254 & 0.4643 & 0.0864 & 0.0931 & 0.1007 & 0.1007 & 0.4777 & 0.4893 & 0.4950 & 0.4973 \\
			& CORR & 0.9843 & 0.9690 & 0.9467 & 0.8870 & 0.9283 & 0.9135 & 0.9077 & 0.9119 & 0.8721 & 0.8690 & 0.8614 & 0.8588 \\ 
			\hline
			TPA-LSTM & RSE & 0.1803 & 0.2347 & 0.3234 & 0.4389 & 0.0823 & 0.0916 & 0.0964 & 0.1006 & 0.4487 & 0.4658 & 0.4641 & 0.4765 \\
			& CORR & 0.9850 & 0.9742 & 0.9487 & \textbf{0.9081} & 0.9439 & 0.9337 & 0.9250 & 0.9133 & 0.8812 & 0.8717 & 0.8717 & 0.8629 \\ \hline
			MTGNN & RSE & 0.1778 & 0.2348 & 0.3109 & 0.4270 & 0.0745 & 0.0878 & 0.0916 & 0.0953 & 0.4162 & 0.4754 & 0.4461 & 0.4535 \\
			& CORR & 0.9852 & 0.9726 & 0.9509 & 0.9031 & 0.9474 & 0.9316 & 0.9278 & 0.9234 & 0.8960 & 0.8667 & 0.8794 & 0.8810 \\ 
			\hline
			HyDCNN & RSE & 0.1806 & 0.2335 & 0.3094 & 0.4225 & 0.0832 & 0.0898 & 0.0921 & 0.0940 &0.4198 & 0.4290 & 0.4352 & 0.4423 \\
			& CORR & 0.9865 &0.9747 & 0.9515 & 0.9096 & 0.9354 & 0.9329 & 0.9285 & 0.9264 & 0.8915 & 0.8855 & 0.8858 & \textbf{0.8819} \\ 
			\hline
			SDLGNN-Corr & RSE & 0.1806 & 0.2378 & 0.3042 & 0.4173 & 0.0737 & 0.0841 & 0.0923 & 0.0971 & 0.4227 & 0.4378 & 0.4576 & 0.4579 \\
			& CORR & 0.9848 & 0.9722 & 0.9534 & 0.9067 & 0.9475 & 0.9346 & 0.9263 & 0.9227 & 0.8937 & 0.8846 & 0.8746 & 0.8784 \\ 
			\hline
			SDLGNN & RSE & 0.1720 & 0.2249 & 0.3024 & 0.4184 & 0.0726 & 0.0820 & 0.0896 & 0.0947 &0.4053 & 0.4209 & 0.4313 & 0.4444 \\
			& CORR & 0.9864 &0.9757 & 0.9547 & 0.9051 & 0.9502 & 0.9384 & 0.9304 & 0.9257 & 0.9017 & 0.8925 & 0.8868 & 0.8801 \\ 
			\hline
			ADLGNN & RSE & \textbf{0.1708} & \textbf{0.2188} &  \textbf{0.2897} & \textbf{0.4128} &\textbf{0.0719} & \textbf{0.0809} & \textbf{0.0887} & \textbf{0.0930} &\textbf{0.4047} & \textbf{0.4201} &\textbf{0.4299 }& \textbf{0.4416} \\
			& CORR & \textbf{0.9866} & \textbf{0.9768} & \textbf{0.9551} &0.9060 & \textbf{0.9506} & \textbf{0.9386} & \textbf{0.9312} & \textbf{0.9294} & \textbf{0.9028} &\textbf{0.8928} & \textbf{0.8876} & 0.8818 \\
			
			\hline
		\end{tabular}%
	}
\end{table}

\subsection{Experimental Setup}
\label{hyperparameters-setup}
For consistency in evaluation, we follow the literature of the state-of-the-art models we compare against. The datasets are divided into training set (60\%), validation set (20\%), and test set (20\%) in a sequential manner \cite{lai_modeling_2018}. 

Three major techniques are used for model training:
\begin{itemize}
	\item \textbf{Increase batch size on the plateau: }The model takes in the two hyper-parameters initial batch size and maximum batch size.
	The dataset is batched at the initial batch size when the training starts and every time a plateau in validation loss is observed over 3 epochs, the batch size is increased by a factor of 2 until it reaches the maximum batch size. 
	
	\item \textbf{Reduce learning rate on the plateau: }This function is active only after the maximum batch size is reached. The learning rate is reduced by a factor of 0.75 every time the validation loss plateaus for over 3 epochs.
	
	\item \textbf{Early stopping: }The training stops when there is no improvement observed in the validation metrics for over 10 epochs.

\end{itemize} 
The hyper-parameters are selected following Lai et al.~\cite{lai_modeling_2018} except for the learning rate which is initially set to a larger value of 0.003, due to the use of larger batch sizes and dynamically adjusted by monitoring the validation loss. The initial batch size is set to be 4 and the maximum batch size is set to 32 except for the traffic dataset which uses a maximum batch size of 16 (we use a low maximum batch size due to size of the dataset and the associated hardware resource limitations).  We use L2 regularization with a penalty of $1\times10^{-4}$ and dropout with 0.3 to prevent overfitting. The depth $k$ of the graph convolution module is set to 2, the $\beta$ hyper-parameter is set to 0.05, $S$ is set to 20, and the number of spatio-temporal blocks is set to 5 following  Wu et al.~\cite{wu2020connecting}. The kernel size for convolutional attention is set to 6. The input window size is set to 168 following Lai et al.~\cite{lai_modeling_2018} and Wu et al.~\cite{wu2020connecting}.

\begin{table}[!htb]
	\caption{Percentage decrease in RSE achieved by ADLGNN over the best from the benchmark models.
	}
	\label{improv-table}
	\centering
		\begin{tabular}{@{}lccccc@{}}
			\hline
			\multicolumn{1}{l}{} & \multicolumn{4}{c}{Horizon} \\ \hline
			Dataset  & 3 & 6 & 12 & 24 & Overall mean \\ \hline
			Solar energy & 3.94\% & 6.3\% & 6.37\% & 2.3\% & \textbf{4.73\%} \\
			
			Electricity  & 3.49\% & 7.86\% & 3.17\% & 1.06\% & \textbf{3.9\%} \\

			Traffic  & 2.76\% & 2.07\% & 1.22\% & 0.16\% & \textbf{1.55\%} \\

			\hline
			Overall mean  & \textbf{3.4\%} & \textbf{5.41\%} & \textbf{3.59\%} & \textbf{1.17\%} &\textbf{3.39\%} \\
		
			\hline
		\end{tabular}%
\end{table}

\subsection{Comparison with State-Of-the-Art Methods}

We perform evaluations for our proposed model Adaptive Dependency Learning Neural Network (ADLNN), two static graph variants of our proposed model, namely Static Dependency Learning Neural Network (SDLNN) and SDLNN-Corr, the state-of-the-art multivariate forecasting models LSTNet \cite{lai_modeling_2018}, TPA-LSTM \cite{tpa_attention_Shih2019}, MTGNN \cite{wu2020connecting}, HyDCNN \cite{li2021modeling},  and the baseline models AR and VAR-MLP. These models are outlined as follows:

\begin{itemize}
    \item \textbf{AR:} An auto-regressive linear model.
	\item \textbf{VAR-MLP:} A composite model of a linear VAR and an MLP \cite{hybrid_arima_mlp03}. 
	\item \textbf{GP:} A Gaussian Process Model \cite{Roberts2013GaussianPF}. 
	\item \textbf{RNN-GRU:} An RNN model with gated recurrent units (GRU).
	\item \textbf{LSTNet:} A neural network model with temporal convolution layers in recurrent fashion with skip connections \cite{lai_modeling_2018}.
	\item \textbf{TPA-LSTM:} An RNN model that uses temporal attention \cite{tpa_attention_Shih2019}.
	\item \textbf{MTGNN:} This GNN model constructs the graph from data. To the best of our knowledge, this is the only work in GNN for multivariate forecasting that constructs graphs from the data \cite{wu2020connecting}.
	\item \textbf{HyDCNN:} This model uses a position-aware fully dilated CNNs to model the non-linear patterns followed by an autoregressive model which models the sequential linear dependencies \cite{li2021modeling}.
	\item \textbf{SDLGNN-Corr:} A variant of our proposed model, which uses pairwise Pearson's correlation to obtain the adjacency matrix which is fed to the graph convolutions as input.
	\item \textbf{SDLGNN:} A static variant of our proposed model, where we use the output from a static graph structure extraction module mentioned in Figure \ref{fig:AdjMat} as adjacency matrix which is fed to the graph convolutions as input.
	\item \textbf{ADLGNN:} Our proposed model.

\end{itemize}

\subsection{Additional experiments}
We perform some additional experiments on a much larger electricity load dataset \cite{jensen2017re} recorded at 1494 nodes across Europe with a series length of 26304. This experiment compares our method against MTGNN as the source code for HyDCNN is not available publicly. We also record the runtime for these experiments to compare the time complexity of both these methods. The experiments use the same hyperparameters listed in section \ref{hyperparameters-setup}. The Experiments were all conducted on a machine with 20vCPUs, 200GiB RAM, 512TiB SSD, and 1 $\times $ A100 (40GB SXM4) GPU.

\begin{table}[!htb]
	\caption{ADLGNN vs MTGNN on European Electricity Load Dataset \cite{jensen2017re}.
	}
	\label{extra-table}
	\centering
	\resizebox{9.5cm}{!}{%
		\begin{tabular}{@{}llcccc@{}}
			\hline
			\multicolumn{1}{l}{} & \multicolumn{4}{c}{Horizon} \\ \hline
			Model & Metric  & 3 & 6 & 12 & 24 \\ \hline
			MTGNN & RSE & 0.0501 & 0.0613 & 0.0742 & 0.0692 \\
                 & CORR & 0.9544 & 0.9366 & 0.9237 & 0.9346 \\
                 & RunTime(Minutes) & 243.65 & 243.32 & 243.72 & 242.99  \\ \hline
			
			ADLGNN &RSE & \textbf{0.0386} & \textbf{0.0499} & \textbf{0.0641} & \textbf{0.0681}  \\
                & CORR & \textbf{0.9850} & \textbf{0.9754} & \textbf{0.9623} & \textbf{0.9571}  \\
                 & RunTime(Minutes) & \textbf{237.71} & \textbf{237.80} & \textbf{237.94} & \textbf{237.76}  \\
			
			\hline
		\end{tabular}%
	}%
\end{table}

Table \ref{extra-table} shows that ADLGNN is 2.30877\% faster than MTGNN while decreasing the RSE by 13.383\%.

\subsection{Results}

\begin{figure*}[htb]
	\centering
	\includegraphics[width=\textwidth]{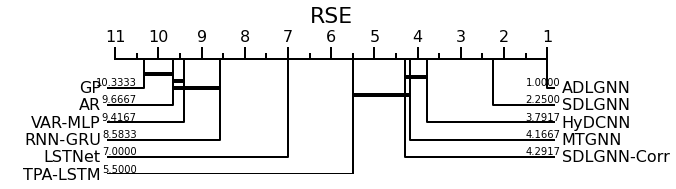}
	\caption{The critical difference diagram shows the mean ranks of each model we considered for
our experiments.}
	\label{fig:cdDia}
\end{figure*}

\begin{table}[!htb]
	\caption{Percentage change in RSE comparing models that use data driven graph construction methods (ablation study).
	}
	\label{ablation-table}
	\centering
	\resizebox{12cm}{!}{%
		\begin{tabular}{@{}lll|cccc|c@{}}
			\hline
			\multicolumn{3}{l|}{} & \multicolumn{4}{c|}{Horizon} \\ \hline
			Dataset & Initial Model & New Model & 3 & 6 & 12 & 24 & Overall mean \\ \hline
			Solar energy 
			& MTGNN & SDLGNN-Corr & -1.57\% & -1.28\% & 2.16\% & 2.27\%  & 0.395\%  \\  
			& SDLGNN-Corr & SDLGNN & 4.76\% & 2.71\% & 0.59\% & -0.26\%  & 1.95\%  \\  
			& SDLGNN & ADLGNN & 0.7\% & 2.71\% & 4.2\% & 1.34\%  & 2.24\%  \\ \
			& MTGNN & ADLGNN & 3.94\% & 6.81\% & 6.82\% & 3.33\%  & 5.23\%  \\ \hline
			
			Electricity  
			
			& MTGNN & SDLGNN-Corr & 1.07\% & 4.21\% & -0.76\% & -1.88\%  & 0.66\%  \\  
			& SDLGNN-Corr & SDLGNN & 1.49\% & 2.5\% & 2.93\% & 2.47\%  & 2.35\%  \\  
			
			& SDLGNN & ADLGNN & 0.96\% & 1.34\% & 1\% & 1.8\%  & 1.28\%  \\ \
			
			& MTGNN & ADLGNN & 3.49\% & 7.86\% & 3.17\% & 2.41\% & 4.17\%  \\ \hline

			Traffic 
			
			& MTGNN & SDLGNN-Corr & -1.56\% & 7.91\% & -2.6\% & -0.97\%  & 0.7\%  \\
			& SDLGNN-Corr & SDLGNN & 4.12\% & 3.91\% & 6.1\% & 2.95\%  & 4.27\%  \\  
		
			& SDLGNN & ADLGNN & 0.15\% & 0.19\% & 0.32\% & 0.63\%  & 0.32\%  \\ 
			& MTGNN & ADLGNN & 2.76\% & 11.63\% & 3.63\% & 2.62\% & 5.16\%  \\ \hline

		\end{tabular}%
	}
\end{table}

In figure \ref{fig:cdDia}, the critical difference diagram compares the accuracy of the models considered in our experiments. The lower the rank, the better the model is at producing more accurate forecasts. The horizontal lines in the diagram connecting the models indicate that the connected models are not significantly different from each other. Figure \ref{fig:cdDia}, shows that there exists a significant difference in accuracy recorded in our experiments between the ADLGNN and the other models. 

Table~\ref{results-table} shows the experimental results of the proposed models along with other existing methods. From the results, we can observe that our proposed models SDLGNN and ADLGNN have outperformed all other models. 

Table \ref{improv-table} shows the percentage decrease in error metric RSE that the proposed models achieved over the best among the baseline and benchmark models. For all the tasks, our proposed models surpassed the performance of benchmark methods. The best of all the proposed model variants is ADLGNN which has recorded the lowest RSE across all tasks. From Table \ref{ablation-table}, we can see that introducing statistically initialised dynamic graphs in ADLGNN improved forecasting accuracy by 4.9\% over MTGNN which is an existing state-of-the-art GNN model that similarly uses a trainable graph. 

It can be observed that we have recorded maximum improvement over the traffic \& solar datasets in relation to the other. The nature of these datasets makes dependency learning more important. The time series in the traffic dataset are more complex and have strong inter-series dependencies. Moreover, the performance of the MTGNN model seems to be degrading with an increase in the number of variables, which ADLGNN and SDLGNN are able to overcome.

\section{Conclusions}

We have proposed a hybrid approach for combining neural networks and statistical structure learning models to learn the dependencies across multivariate data and construct a dynamically changing dependency graph. Therewith, we are able to enable the use of GNNs even where no pre-defined graph is available. The statistical structure modeling together with neural networks provides an efficient approach as it brings causal semantics to determine the dependencies across the series. We have been able to demonstrate significantly improved performance on several real-world datasets, outperforming all the competitor methods.

\section*{Acknowledgment}

This research was supported by the Australian Research Council under grant DE190100045, a Facebook Statistics for Improving Insights and Decisions research award, Monash University Graduate Research funding, and the MASSIVE - High performance computing facility, Australia.

\bibliography{mybibfile}

\begin{thebibliography}{50}
\expandafter\ifx\csname natexlab\endcsname\relax\def\natexlab#1{#1}\fi
\providecommand{\url}[1]{\texttt{#1}}
\providecommand{\href}[2]{#2}
\providecommand{\path}[1]{#1}
\providecommand{\DOIprefix}{doi:}
\providecommand{\ArXivprefix}{arXiv:}
\providecommand{\URLprefix}{URL: }
\providecommand{\Pubmedprefix}{pmid:}
\providecommand{\doi}[1]{\href{http://dx.doi.org/#1}{\path{#1}}}
\providecommand{\Pubmed}[1]{\href{pmid:#1}{\path{#1}}}
\providecommand{\bibinfo}[2]{#2}
\ifx\xfnm\relax \def\xfnm[#1]{\unskip,\space#1}\fi
\bibitem[{Ali et~al.(2021)Ali, Zhu and Zakarya}]{ali2021exploiting}
\bibinfo{author}{Ali, A.}, \bibinfo{author}{Zhu, Y.}, \bibinfo{author}{Zakarya, M.}, \bibinfo{year}{2021}.
\newblock \bibinfo{title}{Exploiting dynamic spatio-temporal correlations for citywide traffic flow prediction using attention based neural networks}.
\newblock \bibinfo{journal}{Information Sciences} \bibinfo{volume}{577}, \bibinfo{pages}{852--870}.
\bibitem[{Bahdanau et~al.(2015)Bahdanau, Cho and Bengio}]{Bahdanau2015NeuralMT}
\bibinfo{author}{Bahdanau, D.}, \bibinfo{author}{Cho, K.}, \bibinfo{author}{Bengio, Y.}, \bibinfo{year}{2015}.
\newblock \bibinfo{title}{Neural machine translation by jointly learning to align and translate}.
\newblock \bibinfo{note}{ICLR 2015}.
\bibitem[{Bai et~al.(2018)Bai, Kolter and Koltun}]{bai2018empirical}
\bibinfo{author}{Bai, S.}, \bibinfo{author}{Kolter, J.Z.}, \bibinfo{author}{Koltun, V.}, \bibinfo{year}{2018}.
\newblock \bibinfo{title}{An empirical evaluation of generic convolutional and recurrent networks for sequence modeling}.
\newblock \bibinfo{journal}{arXiv preprint arXiv:1803.01271} .
\bibitem[{Bandara et~al.(2021)Bandara, Hewamalage, Liu, Kang and Bergmeir}]{bandara2021improving}
\bibinfo{author}{Bandara, K.}, \bibinfo{author}{Hewamalage, H.}, \bibinfo{author}{Liu, Y.H.}, \bibinfo{author}{Kang, Y.}, \bibinfo{author}{Bergmeir, C.}, \bibinfo{year}{2021}.
\newblock \bibinfo{title}{Improving the accuracy of global forecasting models using time series data augmentation}.
\newblock \bibinfo{journal}{Pattern Recognition} , \bibinfo{pages}{108148}.
\bibitem[{Box and Jenkins(1976)}]{boxjen76}
\bibinfo{author}{Box, G.}, \bibinfo{author}{Jenkins, G.M.}, \bibinfo{year}{1976}.
\newblock \bibinfo{title}{Time Series Analysis: Forecasting and Control}.
\newblock \bibinfo{publisher}{Holden-Day}.
\bibitem[{Chen et~al.(2021)Chen, Jiang, Zhang and Chen}]{chen2021novel}
\bibinfo{author}{Chen, W.}, \bibinfo{author}{Jiang, M.}, \bibinfo{author}{Zhang, W.G.}, \bibinfo{author}{Chen, Z.}, \bibinfo{year}{2021}.
\newblock \bibinfo{title}{A novel graph convolutional feature based convolutional neural network for stock trend prediction}.
\newblock \bibinfo{journal}{Information Sciences} \bibinfo{volume}{556}, \bibinfo{pages}{67--94}.
\bibitem[{Fan et~al.(2019)Fan, Zhang, Pan, Li, Zhang, Yuan, Wu, Wang, Pei and Huang}]{fan_multi_horizon_2019}
\bibinfo{author}{Fan, C.}, \bibinfo{author}{Zhang, Y.}, \bibinfo{author}{Pan, Y.}, \bibinfo{author}{Li, X.}, \bibinfo{author}{Zhang, C.}, \bibinfo{author}{Yuan, R.}, \bibinfo{author}{Wu, D.}, \bibinfo{author}{Wang, W.}, \bibinfo{author}{Pei, J.}, \bibinfo{author}{Huang, H.}, \bibinfo{year}{2019}.
\newblock \bibinfo{title}{Multi-horizon time series forecasting with temporal attention learning}, in: \bibinfo{booktitle}{Proceedings of the 25th ACM SIGKDD}, p. \bibinfo{pages}{2527–2535}.
\bibitem[{Friedman et~al.(2008)Friedman, Hastie and Tibshirani}]{friedman2008sparse}
\bibinfo{author}{Friedman, J.}, \bibinfo{author}{Hastie, T.}, \bibinfo{author}{Tibshirani, R.}, \bibinfo{year}{2008}.
\newblock \bibinfo{title}{Sparse inverse covariance estimation with the graphical lasso}.
\newblock \bibinfo{journal}{Biostatistics} \bibinfo{volume}{9}, \bibinfo{pages}{432--441}.
\bibitem[{Guan et~al.(2022)Guan, Iyer and Kim}]{guan2022dynagraph}
\bibinfo{author}{Guan, M.}, \bibinfo{author}{Iyer, A.P.}, \bibinfo{author}{Kim, T.}, \bibinfo{year}{2022}.
\newblock \bibinfo{title}{Dynagraph: dynamic graph neural networks at scale}, in: \bibinfo{booktitle}{Proceedings of the 5th ACM SIGMOD Joint International Workshop on Graph Data Management Experiences \& Systems (GRADES) and Network Data Analytics (NDA)}, pp. \bibinfo{pages}{1--10}.
\bibitem[{Guo et~al.(2019)Guo, Lin and Antulov{-}Fantulin}]{guo_exploring_2019}
\bibinfo{author}{Guo, T.}, \bibinfo{author}{Lin, T.}, \bibinfo{author}{Antulov{-}Fantulin, N.}, \bibinfo{year}{2019}.
\newblock \bibinfo{title}{Exploring interpretable {LSTM} neural networks over multi-variable data} \bibinfo{volume}{97}, \bibinfo{pages}{2494--2504}.
\bibitem[{Hartle et~al.(2020)Hartle, Klein, McCabe, Daniels, St-Onge, Murphy and H{\'e}bert-Dufresne}]{hartle2020network}
\bibinfo{author}{Hartle, H.}, \bibinfo{author}{Klein, B.}, \bibinfo{author}{McCabe, S.}, \bibinfo{author}{Daniels, A.}, \bibinfo{author}{St-Onge, G.}, \bibinfo{author}{Murphy, C.}, \bibinfo{author}{H{\'e}bert-Dufresne, L.}, \bibinfo{year}{2020}.
\newblock \bibinfo{title}{Network comparison and the within-ensemble graph distance}.
\newblock \bibinfo{journal}{Proceedings of the Royal Society A} \bibinfo{volume}{476}, \bibinfo{pages}{20190744}.
\bibitem[{Hochreiter and Schmidhuber(1997)}]{hochreiter1997long}
\bibinfo{author}{Hochreiter, S.}, \bibinfo{author}{Schmidhuber, J.}, \bibinfo{year}{1997}.
\newblock \bibinfo{title}{Long short-term memory}.
\newblock \bibinfo{journal}{Neural computation} \bibinfo{volume}{9}, \bibinfo{pages}{1735--1780}.
\bibitem[{Huang et~al.(2019)Huang, Wang, Wu and Tang}]{dual_self_attn_2019}
\bibinfo{author}{Huang, S.}, \bibinfo{author}{Wang, D.}, \bibinfo{author}{Wu, X.}, \bibinfo{author}{Tang, A.}, \bibinfo{year}{2019}.
\newblock \bibinfo{title}{Dsanet: Dual self-attention network for multivariate time series forecasting}, in: \bibinfo{booktitle}{ACM CIKM}, pp. \bibinfo{pages}{2129--2132}.
\bibitem[{Huang et~al.(2022)Huang, Zhang, Wang and Yin}]{huang2022gan}
\bibinfo{author}{Huang, Z.}, \bibinfo{author}{Zhang, W.}, \bibinfo{author}{Wang, D.}, \bibinfo{author}{Yin, Y.}, \bibinfo{year}{2022}.
\newblock \bibinfo{title}{A {GAN} framework-based dynamic multi-graph convolutional network for origin-destination-based ride-hailing demand prediction}.
\newblock \bibinfo{journal}{Information Sciences} \bibinfo{volume}{601}, \bibinfo{pages}{129--146}.
\bibitem[{Jensen and Pinson(2017)}]{jensen2017re}
\bibinfo{author}{Jensen, T.V.}, \bibinfo{author}{Pinson, P.}, \bibinfo{year}{2017}.
\newblock \bibinfo{title}{Re-europe, a large-scale dataset for modeling a highly renewable european electricity system}.
\newblock \bibinfo{journal}{Scientific data} \bibinfo{volume}{4}, \bibinfo{pages}{1--18}.
\bibitem[{Jiang and Luo(2021)}]{jiang2021graph}
\bibinfo{author}{Jiang, W.}, \bibinfo{author}{Luo, J.}, \bibinfo{year}{2021}.
\newblock \bibinfo{title}{Graph neural network for traffic forecasting: A survey}.
\newblock \bibinfo{journal}{arXiv preprint arXiv:2101.11174} .
\bibitem[{Jin et~al.(2019)Jin, Liu, Li, He and Zhang}]{jin2019graph}
\bibinfo{author}{Jin, D.}, \bibinfo{author}{Liu, Z.}, \bibinfo{author}{Li, W.}, \bibinfo{author}{He, D.}, \bibinfo{author}{Zhang, W.}, \bibinfo{year}{2019}.
\newblock \bibinfo{title}{Graph convolutional networks meet markov random fields: Semi-supervised community detection in attribute networks}, in: \bibinfo{booktitle}{Proceedings of the AAAI conference on artificial intelligence}, pp. \bibinfo{pages}{152--159}.
\bibitem[{Klicpera et~al.(2018)Klicpera, Bojchevski and G{\"u}nnemann}]{klicpera2018predict}
\bibinfo{author}{Klicpera, J.}, \bibinfo{author}{Bojchevski, A.}, \bibinfo{author}{G{\"u}nnemann, S.}, \bibinfo{year}{2018}.
\newblock \bibinfo{title}{Predict then propagate: Graph neural networks meet personalized pagerank}.
\newblock \bibinfo{journal}{arXiv preprint arXiv:1810.05997} .
\bibitem[{Klicpera et~al.(2019)Klicpera, Bojchevski and Günnemann}]{klicpera2018combining}
\bibinfo{author}{Klicpera, J.}, \bibinfo{author}{Bojchevski, A.}, \bibinfo{author}{Günnemann, S.}, \bibinfo{year}{2019}.
\newblock \bibinfo{title}{Combining neural networks with personalized pagerank for classification on graphs}, in: \bibinfo{booktitle}{ICLR}.
\bibitem[{Kolassa(2016)}]{kolassa2016evaluating}
\bibinfo{author}{Kolassa, S.}, \bibinfo{year}{2016}.
\newblock \bibinfo{title}{Evaluating predictive count data distributions in retail sales forecasting}.
\newblock \bibinfo{journal}{International Journal of Forecasting} \bibinfo{volume}{32}, \bibinfo{pages}{788--803}.
\bibitem[{Lai et~al.()Lai, Chang, Yang and Liu}]{lai_modeling_2018}
\bibinfo{author}{Lai, G.}, \bibinfo{author}{Chang, W.C.}, \bibinfo{author}{Yang, Y.}, \bibinfo{author}{Liu, H.}, .
\newblock \bibinfo{title}{Modeling long- and short-term temporal patterns with deep neural networks}, in: \bibinfo{booktitle}{ACM SIGIR}, pp. \bibinfo{pages}{95--104}.
\bibitem[{Li et~al.(2019)Li, Jin, Xuan, Zhou, Chen, Wang and Yan}]{li2019enhancing}
\bibinfo{author}{Li, S.}, \bibinfo{author}{Jin, X.}, \bibinfo{author}{Xuan, Y.}, \bibinfo{author}{Zhou, X.}, \bibinfo{author}{Chen, W.}, \bibinfo{author}{Wang, Y.X.}, \bibinfo{author}{Yan, X.}, \bibinfo{year}{2019}.
\newblock \bibinfo{title}{Enhancing the locality and breaking the memory bottleneck of transformer on time series forecasting}.
\newblock \bibinfo{journal}{Advances in Neural Information Processing Systems} \bibinfo{volume}{32}, \bibinfo{pages}{5243--5253}.
\bibitem[{Li et~al.(2020)Li, Zhang, Wu, Liu, Wang and Philip}]{li2020dynamic}
\bibinfo{author}{Li, X.}, \bibinfo{author}{Zhang, M.}, \bibinfo{author}{Wu, S.}, \bibinfo{author}{Liu, Z.}, \bibinfo{author}{Wang, L.}, \bibinfo{author}{Philip, S.Y.}, \bibinfo{year}{2020}.
\newblock \bibinfo{title}{Dynamic graph collaborative filtering}, in: \bibinfo{booktitle}{2020 IEEE International Conference on Data Mining (ICDM)}, \bibinfo{organization}{IEEE}. pp. \bibinfo{pages}{322--331}.
\bibitem[{Li et~al.(2021)Li, Li, Chen, Zhou, Zeng and Li}]{li2021modeling}
\bibinfo{author}{Li, Y.}, \bibinfo{author}{Li, K.}, \bibinfo{author}{Chen, C.}, \bibinfo{author}{Zhou, X.}, \bibinfo{author}{Zeng, Z.}, \bibinfo{author}{Li, K.}, \bibinfo{year}{2021}.
\newblock \bibinfo{title}{Modeling temporal patterns with dilated convolutions for time-series forecasting}.
\newblock \bibinfo{journal}{ACM Transactions on Knowledge Discovery from Data (TKDD)} \bibinfo{volume}{16}, \bibinfo{pages}{1--22}.
\bibitem[{Li et~al.(2018)Li, Nie, Chang, Yang, Zhang and Sebe}]{li2018dynamic}
\bibinfo{author}{Li, Z.}, \bibinfo{author}{Nie, F.}, \bibinfo{author}{Chang, X.}, \bibinfo{author}{Yang, Y.}, \bibinfo{author}{Zhang, C.}, \bibinfo{author}{Sebe, N.}, \bibinfo{year}{2018}.
\newblock \bibinfo{title}{Dynamic affinity graph construction for spectral clustering using multiple features}.
\newblock \bibinfo{journal}{IEEE transactions on neural networks and learning systems} \bibinfo{volume}{29}, \bibinfo{pages}{6323--6332}.
\bibitem[{Lim et~al.()Lim, Arik, Loeff and Pfister}]{lim_temporal_2019}
\bibinfo{author}{Lim, B.}, \bibinfo{author}{Arik, S.O.}, \bibinfo{author}{Loeff, N.}, \bibinfo{author}{Pfister, T.}, .
\newblock \bibinfo{title}{Temporal fusion transformers for interpretable multi-horizon time series forecasting} \href{http://arxiv.org/abs/1912.09363}{{\tt arXiv:1912.09363}}.
\bibitem[{Lizier(2014)}]{lizier2014jidt}
\bibinfo{author}{Lizier, J.T.}, \bibinfo{year}{2014}.
\newblock \bibinfo{title}{Jidt: An information-theoretic toolkit for studying the dynamics of complex systems}.
\newblock \bibinfo{journal}{Frontiers in Robotics and AI} \bibinfo{volume}{1}, \bibinfo{pages}{11}.
\bibitem[{Luo et~al.(2017)Luo, Chang, Nie, Yang, Hauptmann and Zheng}]{luo2017adaptive}
\bibinfo{author}{Luo, M.}, \bibinfo{author}{Chang, X.}, \bibinfo{author}{Nie, L.}, \bibinfo{author}{Yang, Y.}, \bibinfo{author}{Hauptmann, A.G.}, \bibinfo{author}{Zheng, Q.}, \bibinfo{year}{2017}.
\newblock \bibinfo{title}{An adaptive semisupervised feature analysis for video semantic recognition}.
\newblock \bibinfo{journal}{IEEE transactions on cybernetics} \bibinfo{volume}{48}, \bibinfo{pages}{648--660}.
\bibitem[{Luong et~al.(2015)Luong, Pham and Manning}]{luong-etal-2015-effective}
\bibinfo{author}{Luong, T.}, \bibinfo{author}{Pham, H.}, \bibinfo{author}{Manning, C.D.}, \bibinfo{year}{2015}.
\newblock \bibinfo{title}{Effective approaches to attention-based neural machine translation}, in: \bibinfo{booktitle}{Proceedings of the 2015 Conference on Empirical Methods in Natural Language Processing}, \bibinfo{publisher}{Association for Computational Linguistics}, \bibinfo{address}{Lisbon, Portugal}. pp. \bibinfo{pages}{1412--1421}.
\bibitem[{Mantegna(1999)}]{mantegna1999hierarchical}
\bibinfo{author}{Mantegna, R.N.}, \bibinfo{year}{1999}.
\newblock \bibinfo{title}{Hierarchical structure in financial markets}.
\newblock \bibinfo{journal}{The European Physical Journal B-Condensed Matter and Complex Systems} \bibinfo{volume}{11}, \bibinfo{pages}{193--197}.
\bibitem[{Peng et~al.(2021)Peng, Du, Liu, Liu, Ji, Wang, Zhang and He}]{peng2021dynamic}
\bibinfo{author}{Peng, H.}, \bibinfo{author}{Du, B.}, \bibinfo{author}{Liu, M.}, \bibinfo{author}{Liu, M.}, \bibinfo{author}{Ji, S.}, \bibinfo{author}{Wang, S.}, \bibinfo{author}{Zhang, X.}, \bibinfo{author}{He, L.}, \bibinfo{year}{2021}.
\newblock \bibinfo{title}{Dynamic graph convolutional network for long-term traffic flow prediction with reinforcement learning}.
\newblock \bibinfo{journal}{Information Sciences} \bibinfo{volume}{578}, \bibinfo{pages}{401--416}.
\bibitem[{Peng et~al.(2020)Peng, Wang, Du, Bhuiyan, Ma, Liu, Wang, Yang, Du, Wang et~al.}]{peng2020spatial}
\bibinfo{author}{Peng, H.}, \bibinfo{author}{Wang, H.}, \bibinfo{author}{Du, B.}, \bibinfo{author}{Bhuiyan, M.Z.A.}, \bibinfo{author}{Ma, H.}, \bibinfo{author}{Liu, J.}, \bibinfo{author}{Wang, L.}, \bibinfo{author}{Yang, Z.}, \bibinfo{author}{Du, L.}, \bibinfo{author}{Wang, S.}, et~al., \bibinfo{year}{2020}.
\newblock \bibinfo{title}{Spatial temporal incidence dynamic graph neural networks for traffic flow forecasting}.
\newblock \bibinfo{journal}{Information Sciences} \bibinfo{volume}{521}, \bibinfo{pages}{277--290}.
\bibitem[{Qu et~al.(2019)Qu, Bengio and Tang}]{qu2019gmnn}
\bibinfo{author}{Qu, M.}, \bibinfo{author}{Bengio, Y.}, \bibinfo{author}{Tang, J.}, \bibinfo{year}{2019}.
\newblock \bibinfo{title}{Gmnn: Graph markov neural networks}, in: \bibinfo{booktitle}{International conference on machine learning}, \bibinfo{organization}{PMLR}. pp. \bibinfo{pages}{5241--5250}.
\bibitem[{Qu and Tang(2019)}]{qu2019probabilistic}
\bibinfo{author}{Qu, M.}, \bibinfo{author}{Tang, J.}, \bibinfo{year}{2019}.
\newblock \bibinfo{title}{Probabilistic logic neural networks for reasoning}.
\newblock \bibinfo{journal}{Advances in neural information processing systems} \bibinfo{volume}{32}.
\bibitem[{Roberts et~al.(2013)Roberts, Osborne, Ebden, Reece, Gibson and Aigrain}]{Roberts2013GaussianPF}
\bibinfo{author}{Roberts, S.}, \bibinfo{author}{Osborne, M.}, \bibinfo{author}{Ebden, M.}, \bibinfo{author}{Reece, S.}, \bibinfo{author}{Gibson, N.}, \bibinfo{author}{Aigrain, S.}, \bibinfo{year}{2013}.
\newblock \bibinfo{title}{Gaussian processes for time-series modelling}.
\newblock \bibinfo{journal}{Philosophical Transactions of the Royal Society A: Mathematical, Physical and Engineering Sciences} \bibinfo{volume}{371}.
\bibitem[{Scutari(2009)}]{scutari2009learning}
\bibinfo{author}{Scutari, M.}, \bibinfo{year}{2009}.
\newblock \bibinfo{title}{Learning bayesian networks with the bnlearn r package}.
\newblock \bibinfo{journal}{arXiv preprint arXiv:0908.3817} .
\bibitem[{Shih et~al.(2019)Shih, Sun and Lee}]{tpa_attention_Shih2019}
\bibinfo{author}{Shih, S.Y.}, \bibinfo{author}{Sun, F.K.}, \bibinfo{author}{Lee, H.Y.}, \bibinfo{year}{2019}.
\newblock \bibinfo{title}{Temporal pattern attention for multivariate time series forecasting.} \bibinfo{volume}{108}, \bibinfo{pages}{1421--1441}.
\bibitem[{Sun et~al.(2022)Sun, Wu, Wang and Ye}]{sun2022sequential}
\bibinfo{author}{Sun, Z.}, \bibinfo{author}{Wu, B.}, \bibinfo{author}{Wang, Y.}, \bibinfo{author}{Ye, Y.}, \bibinfo{year}{2022}.
\newblock \bibinfo{title}{Sequential graph collaborative filtering}.
\newblock \bibinfo{journal}{Information Sciences} \bibinfo{volume}{592}, \bibinfo{pages}{244--260}.
\bibitem[{Sutskever et~al.(2014)Sutskever, Vinyals and Le}]{sutskever2014sequence}
\bibinfo{author}{Sutskever, I.}, \bibinfo{author}{Vinyals, O.}, \bibinfo{author}{Le, Q.V.}, \bibinfo{year}{2014}.
\newblock \bibinfo{title}{Sequence to sequence learning with neural networks}, in: \bibinfo{booktitle}{Advances in neural information processing systems}, pp. \bibinfo{pages}{3104--3112}.
\bibitem[{Vaswani et~al.(2017)Vaswani, Shazeer, Parmar, Uszkoreit, Jones, Gomez, Kaiser and Polosukhin}]{vaswani_attention_17}
\bibinfo{author}{Vaswani, A.}, \bibinfo{author}{Shazeer, N.}, \bibinfo{author}{Parmar, N.}, \bibinfo{author}{Uszkoreit, J.}, \bibinfo{author}{Jones, L.}, \bibinfo{author}{Gomez, A.N.}, \bibinfo{author}{Kaiser, L.u.}, \bibinfo{author}{Polosukhin, I.}, \bibinfo{year}{2017}.
\newblock \bibinfo{title}{Attention is all you need}, in: \bibinfo{booktitle}{Advances in Neural Information Processing Systems 30}. \bibinfo{publisher}{Curran Associates, Inc.}, pp. \bibinfo{pages}{5998--6008}.
\bibitem[{{Wu} et~al.(2020){Wu}, {Pan}, {Chen}, {Long}, {Zhang} and {Yu}}]{gnn_review_2019}
\bibinfo{author}{{Wu}, Z.}, \bibinfo{author}{{Pan}, S.}, \bibinfo{author}{{Chen}, F.}, \bibinfo{author}{{Long}, G.}, \bibinfo{author}{{Zhang}, C.}, \bibinfo{author}{{Yu}, P.S.}, \bibinfo{year}{2020}.
\newblock \bibinfo{title}{A comprehensive survey on graph neural networks}.
\newblock \bibinfo{journal}{IEEE TNNLS} , \bibinfo{pages}{1--21}.
\bibitem[{Wu et~al.(2020)Wu, Pan, Long, Jiang, Chang and Zhang}]{wu2020connecting}
\bibinfo{author}{Wu, Z.}, \bibinfo{author}{Pan, S.}, \bibinfo{author}{Long, G.}, \bibinfo{author}{Jiang, J.}, \bibinfo{author}{Chang, X.}, \bibinfo{author}{Zhang, C.}, \bibinfo{year}{2020}.
\newblock \bibinfo{title}{Connecting the dots: Multivariate time series forecasting with graph neural networks}, in: \bibinfo{booktitle}{Proceedings of the 26th ACM SIGKDD}.
\bibitem[{Xu et~al.(2015)Xu, Ba, Kiros, Cho, Courville, Salakhudinov, Zemel and Bengio}]{pmlr-v37-xuc15}
\bibinfo{author}{Xu, K.}, \bibinfo{author}{Ba, J.}, \bibinfo{author}{Kiros, R.}, \bibinfo{author}{Cho, K.}, \bibinfo{author}{Courville, A.}, \bibinfo{author}{Salakhudinov, R.}, \bibinfo{author}{Zemel, R.}, \bibinfo{author}{Bengio, Y.}, \bibinfo{year}{2015}.
\newblock \bibinfo{title}{Show, attend and tell: Neural image caption generation with visual attention}, \bibinfo{publisher}{PMLR}, \bibinfo{address}{Lille, France}. pp. \bibinfo{pages}{2048--2057}.
\bibitem[{Yan et~al.(2018)Yan, Xiong and Lin}]{yan2018spatial}
\bibinfo{author}{Yan, S.}, \bibinfo{author}{Xiong, Y.}, \bibinfo{author}{Lin, D.}, \bibinfo{year}{2018}.
\newblock \bibinfo{title}{Spatial temporal graph convolutional networks for skeleton-based action recognition}.
\newblock \bibinfo{journal}{AAAI} , \bibinfo{pages}{7444--7452}.
\bibitem[{You et~al.(2022)You, Du and Leskovec}]{you2022roland}
\bibinfo{author}{You, J.}, \bibinfo{author}{Du, T.}, \bibinfo{author}{Leskovec, J.}, \bibinfo{year}{2022}.
\newblock \bibinfo{title}{Roland: graph learning framework for dynamic graphs}, in: \bibinfo{booktitle}{Proceedings of the 28th ACM SIGKDD Conference on Knowledge Discovery and Data Mining}, pp. \bibinfo{pages}{2358--2366}.
\bibitem[{Zeng et~al.(2013)Zeng, Alava, Aurell, Hertz and Roudi}]{zeng2013maximum}
\bibinfo{author}{Zeng, H.L.}, \bibinfo{author}{Alava, M.}, \bibinfo{author}{Aurell, E.}, \bibinfo{author}{Hertz, J.}, \bibinfo{author}{Roudi, Y.}, \bibinfo{year}{2013}.
\newblock \bibinfo{title}{Maximum likelihood reconstruction for ising models with asynchronous updates}.
\newblock \bibinfo{journal}{Physical review letters} \bibinfo{volume}{110}, \bibinfo{pages}{210601}.
\bibitem[{Zhang(2003)}]{hybrid_arima_mlp03}
\bibinfo{author}{Zhang, G.P.}, \bibinfo{year}{2003}.
\newblock \bibinfo{title}{Time series forecasting using a hybrid arima and neural network model.}
\newblock \bibinfo{journal}{Neurocomputing} \bibinfo{volume}{50}, \bibinfo{pages}{159--175}.
\bibitem[{Zhang et~al.(2020)Zhang, Chen, Yang, Ramamurthy, Li, Qi and Song}]{zhang2020efficient}
\bibinfo{author}{Zhang, Y.}, \bibinfo{author}{Chen, X.}, \bibinfo{author}{Yang, Y.}, \bibinfo{author}{Ramamurthy, A.}, \bibinfo{author}{Li, B.}, \bibinfo{author}{Qi, Y.}, \bibinfo{author}{Song, L.}, \bibinfo{year}{2020}.
\newblock \bibinfo{title}{Efficient probabilistic logic reasoning with graph neural networks}.
\newblock \bibinfo{journal}{arXiv preprint arXiv:2001.11850} .
\bibitem[{Zhou et~al.(2019)Zhou, Chang, Shi, Shen, Yang and Nie}]{zhou2019person}
\bibinfo{author}{Zhou, R.}, \bibinfo{author}{Chang, X.}, \bibinfo{author}{Shi, L.}, \bibinfo{author}{Shen, Y.D.}, \bibinfo{author}{Yang, Y.}, \bibinfo{author}{Nie, F.}, \bibinfo{year}{2019}.
\newblock \bibinfo{title}{Person reidentification via multi-feature fusion with adaptive graph learning}.
\newblock \bibinfo{journal}{IEEE transactions on neural networks and learning systems} \bibinfo{volume}{31}, \bibinfo{pages}{1592--1601}.
\bibitem[{Zivot and Wang(2003)}]{Zivot2003VectorAM}
\bibinfo{author}{Zivot, E.}, \bibinfo{author}{Wang, J.}, \bibinfo{year}{2003}.
\newblock \bibinfo{title}{Vector autoregressive models for multivariate time series}, in: \bibinfo{booktitle}{Modeling Financial Time Series with S-Plus{\textregistered}}, \bibinfo{publisher}{Springer}. pp. \bibinfo{pages}{369--413}.

\end{thebibliography}

\end{document}